\documentclass[11pt,a4paper]{article}
\usepackage[hyperref]{emnlp2020}
\usepackage{times}
\usepackage{latexsym}

\usepackage{microtype}

\usepackage{graphicx}
\usepackage{booktabs}
\usepackage{multirow}

\definecolor{green1}{RGB}{0, 176, 80}

\aclfinalcopy % Uncomment this line for the final submission
\title{Adapting BERT for Word Sense Disambiguation with \\ Gloss Selection Objective and Example Sentences}

\author{Boon Peng Yap, Andrew Koh, Eng Siong Chng \\
  Nanyang Technological University, Singapore \\
  {\tt \{boonpeng001, andr0081, aseschng\}@ntu.edu.sg} \\}

\date{}

\begin{document}
\maketitle
\begin{abstract}
Domain adaptation or transfer learning using pre-trained language models such as BERT has proven to be an effective approach for many natural language processing tasks. In this work, we propose to formulate word sense disambiguation as a relevance ranking task, and fine-tune BERT on sequence-pair ranking task to select the most probable sense definition given a context sentence and a list of candidate sense definitions. We also introduce a data augmentation technique for WSD using existing example sentences from WordNet. Using the proposed training objective and data augmentation technique, our models are able to achieve state-of-the-art results on the English all-words benchmark datasets.\footnote{Codes and pre-trained models are available at \url{https://github.com/BPYap/BERT-WSD}.}
\end{abstract}

\section{Introduction}

In natural language processing, Word Sense Disambiguation (WSD) refers to the task of identifying the exact sense of an ambiguous word given the context \cite{navigli2009word}. More specifically, WSD associates ambiguous words with predefined senses from an external sense inventory, e.g. WordNet \cite{miller1995wordnet} and BabelNet \cite{navigli2010babelnet}.

% Historically, WSD systems were classified into 2 main categories: knowledge-based and supervised system. Knowledge-based systems rely on manually-curated lexical resources and can be further subdivided into 2 approaches, including similarity-based approaches \cite{lesk1986automatic,banerjee2003extended,chen2014unified} and graph-based approaches \cite{agirre2009personalizing,ponzetto2010knowledge,moro2014entity}. Supervised systems include models trained on corpora manually annotated for WSD, using methods such as support vector machine \cite{iacobacci2016embeddings}, simple linear classifier \cite{zhong2010makes,shen2013coarse} and LSTM-based classifier \cite{yuan2016semi,raganato2017neural,luo2018leveraging}. In terms of performance, supervised WSD systems usually outperform the knowledge-based WSD systems. However, supervised systems tend to suffer from insufficient sense coverage in training corpora which hinders their ability to generalize to unseen words and languages.m

Recent studies in learning contextualized word representations from language models, e.g. ELMo \cite{peters2018deep}, BERT \cite{devlin2019bert} and GPT-2 \cite{radford2019language} attempt to alleviate the issue of insufficient labeled data by first pre-training a language model on a large text corpus through self-supervised learning. The weights from the pre-trained language model can then be fine-tuned on downstream NLP tasks such as question answering and natural language inference. For WSD, pre-trained BERT has been utilized in multiple ways with varying degrees of success. Notably, \citet{huang2019glossbert} proposed GlossBERT, a model based on fine-tuning BERT on sequence-pair binary classification task, and achieved state-of-the-art results in terms of single model performance on several English all-words WSD benchmark datasets.

In this paper, we extend the sequence-pair WSD model and propose a new task objective that can better exploit the inherent relationships within positive and negative sequence pairs. Briefly, our contribution is two-fold: (1) we formulate WSD as gloss selection task, in which the model learns to select the best context-gloss pair from a group of related pairs; (2) we demonstrate how to make use of additional lexical resources, namely the example sentences from WordNet to further improve WSD performance.

We fine-tune BERT using the gloss selection objective on SemCor \cite{miller1994using} plus additional training instances constructed from the WordNet example sentences and evaluate its impact on several commonly used benchmark datasets for English all-words WSD. Experimental results show that the gloss selection objective can indeed improve WSD performance; and using WordNet example sentences as additional training data can offer further performance boost.

\section{Related Work}

BERT \cite{devlin2019bert} is a language representation model based on multi-layer bidirectional Transformer encoder \cite{vaswani2017attention}. Previous experiment results have showed that significant improvement can be achieved in many downstream NLP tasks through fine-tuning BERT on those tasks. Several methods have been proposed to apply BERT for WSD. In this section, we briefly describe two commonly used approaches: feature-based and fine-tuning approach.

\subsection{Feature-based Approaches}
Feature-based WSD systems make use of contextualized word embeddings from BERT as input features for task-specific architectures. \citet{vial2019sense} used the contextual embeddings as inputs in a Transformer-based classifier. They proposed two sense vocabulary compression techniques to reduce the number of output classes by exploiting the semantic relationships between different senses. The Transformer-based classifiers were trained from scratch using the reduced output classes on SemCor and WordNet Gloss Corpus (WNGC). Their ensemble model, which consists of 8 independently trained classifiers achieved state-of-the-art results on the English all-words WSD benchmark datasets.

Besides deep learning-based approach, \citet{loureiro2019language} and \citet{scarlini2020sensembert} construct sense embeddings using the contextual embeddings from BERT. The former generates sense embeddings by averaging the contextual embeddings of sense-annotated tokens taken from SemCor while the latter constructs sense embeddings by concatenating the contextual embeddings of BabelNet definitions with the contextual embeddings of Wikipedia contexts. For WSD, both approaches make use of the constructed sense embeddings in nearest neighbor classification (kNN), in which the simple 1-nearest neighbor approach from \citet{scarlini2020sensembert} showed substantial improvement over the nominal category of the English all-words WSD benchmark datasets.

\subsection{Fine-tuning Approaches}
Fine-tuning WSD systems directly adjust the pre-trained weights on annotated corpora rather than learning new weights from scratch. \citet{du2019using} fine-tuned two separate and independent BERT models simultaneously: one to encode sense-annotated sentences and another one to encode sense definitions from WordNet. The hidden states from the 2 encoders are then concatenated and used to train a multilayer perceptron classifier for WSD.
% Instead of formulating WSD as a multiclass classification problem where each output neuron corresponds to a sense from training corpora, GlossBERT learns to classify whether a sense definition (gloss) entails the meaning of an ambiguous word in a context. This formulation reduces the number of output neurons to only 2 neurons representing a positive and a negative class, which means the model has less parameters to learn. More importantly, it is not constrained to a limited set of senses from the training corpora and therefore it can be scaled to unseen glosses at the inference stage. 

\citet{huang2019glossbert} proposed GlossBERT which fine-tunes BERT on sequence-pair binary classification tasks. The training data consists of context-gloss pairs constructed using annotated sentences from SemCor and sense definitions from WordNet 3.0. Each context-gloss pair contains a sentence from SemCor with a target word to be disambiguated (context) and a candidate sense definition of the target word from WordNet (gloss). During fine-tuning, GlossBERT classifies each context-gloss pair as either positive or negative depending on whether the sense definition corresponds to the correct sense of the target word in the context. Each context-gloss pair is treated as independent training instance and will be shuffled to a random position at the start of each training epoch. At inference stage, the context-gloss pair with the highest output score from the positive neuron among other candidates is chosen as the best answer. 

In this paper, we use similar context-gloss pairs as inputs for our proposed WSD model. However, instead of treating individual context-gloss pair as independent training instance, we group related context-gloss pairs as 1 training instance, i.e. context-gloss pairs with the same context but different candidate glosses are considered as 1 group. Using groups of context-gloss pairs as training data, we formulate WSD as a ranking/selection problem where the most probable sense is ranked first. By processing all related candidate senses in one go, the WSD model will be able to learn better discriminating features between positive and negative context-gloss pairs.

\section{Methodology}
We describe the implementation details of our approaches in this section. When customizing BERT for WSD, we use a linear layer consisting of just 1 neuron in the output layer to compute the relevance score for each context-gloss pair, in contrast to the binary classification layer used in GlossBERT.

Additionally, we also extract example sentences from WordNet 3.0 and use them as additional training data on top of the sense-annotated sentences from SemCor.

\subsection{Gloss Selection Objective}
\begin{figure*}[ht]
  \centering
  \includegraphics[width=0.98\textwidth]{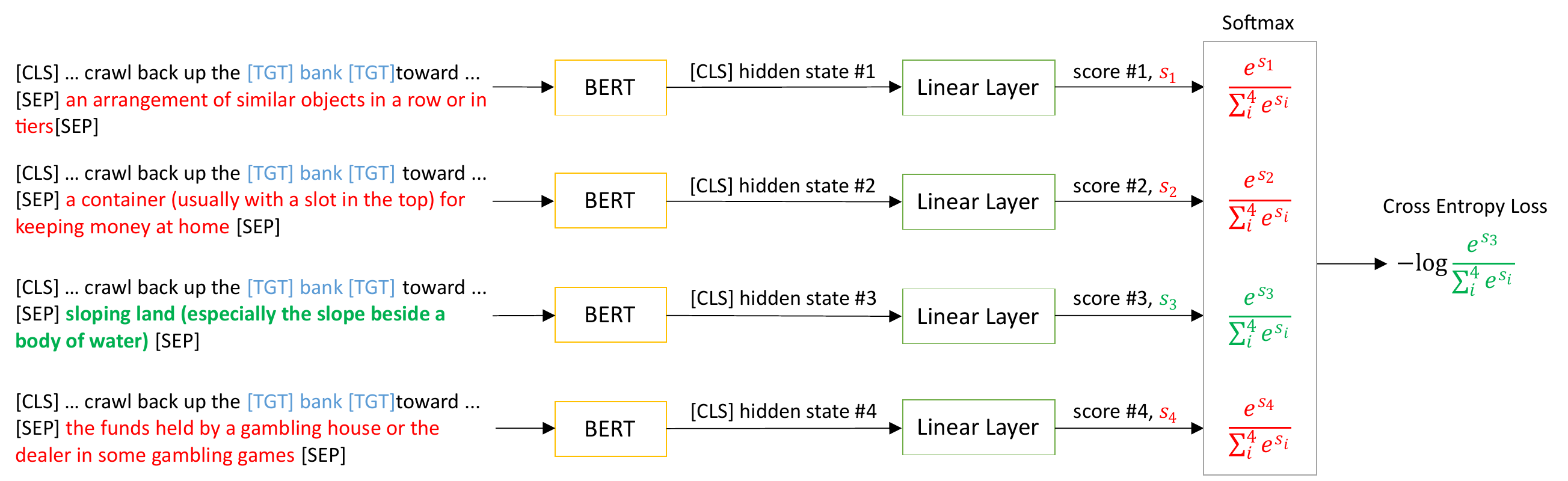}
  \caption{Visualisation of the gloss selection objective when computing the loss value for a training instance. The context “\textit{He turned slowly and began to crawl back up the bank toward the rampart.}" is annotated with the target word “\textit{bank}”. A training instance consists of n context-gloss pairs (n=4 in this case), including 1 positive pair (shown in \textbf{\textcolor{green1}{green}}) and n-1 negative pairs (shown in \textcolor{red}{red}). The order of the context-gloss pairs within each training instance is randomized during the dataset construction step.}
  \label{fig:architecture}
\end{figure*}

Following \citet{huang2019glossbert}, we construct positive and negative context-gloss pairs by combining annotated sentences from SemCor and sense definitions from WordNet 3.0. The positive pair contains a gloss representing the correct sense of the target word while a negative pair contains a negative candidate gloss. Each target word in the contexts is surrounded with two special [TGT] tokens. We group context-gloss pairs with the same context and target word as a single training instance so that they are processed sequentially by the neural network. As illustrated in Figure \ref{fig:architecture}, the output layer takes the hidden states of the [CLS] token from each context-gloss pair as input and calculate the corresponding relevance score. A softmax layer then aggregates the relevance scores from the same group and computes the training loss using cross entropy as loss function. Formally, the gloss selection objective is given as follow:
\begin{equation}
    loss = -\frac{1}{m}\sum_{i=1}^{m}[\sum_{j=1}^{n_{i}}1(y_{i}, j)\mathrm{log}(p_{ij})]
\label{eq:gloss_selection}
\end{equation}
where $m$ is the batch size, $n_{i}$ is number of candidate glosses for the $i$-th training instance, $1(y_{i}, j)$ is the binary indicator if index $j$ is the same as the index of the positive context-gloss pair $y_{i}$, and $p_{ij}$ is the softmax value for the $j$-th candidate sense of $i$-th training instance, computed using the following equation:
\begin{equation}
    p_{ij} = \frac{\mathrm{exp}(Rel(context_{i}, gloss_{ij}))}{\sum_{k}^{n_{i}}\mathrm{exp}(Rel(context_{i}, gloss_{ik}))}
\label{eq:gloss_selection_prob}
\end{equation}
where $Rel(context, gloss)$ denotes the relevance score of a context-gloss pair from the output layer. Similar formulation was presented for web document ranking \cite{huang2013learning} and question-answering natural language inference \cite{liu2019multi}. In the case of WSD, we are only interested in the top-1 context-gloss pair. Hence, during testing, we select the context-gloss pair with the highest relevance score and its corresponding sense as the most probable sense for the target word.

\subsection{Data Augmentation using Example Sentences}
Most synsets in WordNet 3.0 include one or more short sentences illustrating the usage of the synset members (i.e. synonyms). We introduce a relatively straightforward data augmentation technique that combines the example sentences with positive/negative glosses into additional context-gloss pairs. First, example sentences (context) are extracted from each synset and target words are identified via keyword matching and annotated with two [TGT] tokens. Then, context-gloss pairs are constructed by combining the annotated contexts with positive and negative glosses. Using this technique, we were able to obtain 37,596 additional training instances (about 17\%  more training instances).

% \subsubsection{Gloss confusion using stopwords}
% Denoising training objectives have been proven to work better than casual language modeling in the recent years. \cite{devlin2019bert} first introduced the unsupervised masked language modeling objective by corrupting the training input. Random words in the training data are masked with 15\% chance and the model is trained to predict the masked words. Following this, \cite{raffel2019exploring} also used a unsupervised denoising objective whereby random spans in the input are replaced by a sentinel token and the model is trained to predict the missing spans.

% Inspired by the above, we explore confusing the training data by adding a random stopword with a confusion probability $p$ between each word in the gloss for each context-gloss pair in the training data. Stopwords are common and repetitive words which are often filtered out in data preprocessing. The list of stopwords are obtained from NLTK\footnote{https://www.nltk.org/}. We experimented with varying values of $p$ from 0.1 to 0.5 and settled on 0.1 and 0.2 providing the best improvements.

% The rationale between this approach is that stopwords (e.g. are, an, be) often provide less semantic meaning and are there to serve a syntactic purpose. Hence, by adding these words to the glosses, there is effectively almost no change in the semantic meaning of the gloss and the model should still be able to choose the correct answer.

\begin{table*}[ht]
\small
\begin{tabular}{@{}p{0.1cm}p{5.09cm}cccccccccc@{}}
\cmidrule(l){3-12}
                                                                      & \multirow{2}{*}{System}                & Dev           & \multicolumn{4}{c}{Test}                          & \multicolumn{5}{c}{Concatenation of all datasets} \\ \cmidrule(l){3-12} 
                                                                      &                                        & SE07          & SE2            & SE3           & SE13          & SE15                & Noun          & Verb          & Adj           & Adv           & ALL     \\ \hline\hline         
\multirow{3}{*}{\rotatebox[origin=c]{90}{\scriptsize{\textit{KB}}}}   & Most frequent sense baseline           & 54.5          & 65.6           & 66.0          & 63.8          & 67.1                & 67.7          & 49.8          & 73.1          & 80.5          & 65.5    \\
                                                                      & Lesk\textsubscript{ext}+emb            & 56.7          & 63.0           & 63.7          & 66.2          & 64.6                & 70.0          & 51.1          & 51.7          & 80.6          & 64.2    \\
                                                                      & Babelfy                                & 51.6          & 67.0           & 63.5          & 66.4          & 70.3                & 68.9          & 50.7          & 73.2          & 79.8          & 66.4    \\ \hline
\multirow{4}{*}{\rotatebox[origin=c]{90}{\scriptsize{\textit{Sup}}}}  & IMS+emb                                & 62.6          & 72.2           & 70.4          & 65.9          & 71.5                & 71.9          & 56.6          & 75.9          & 84.7          & 70.1    \\
                                                                      & LSTM-LP                                & 63.5          & 73.8           & 71.8          & 69.5          & 72.6                & -             & -             & -             & -             & -       \\
                                                                      & Bi-LSTM                                & -             & 71.1           & 68.4          & 64.8          & 68.3                & 69.5          & 55.9          & 76.2          & 82.4          & 68.4    \\
                                                                      & HCAN                                   & -             & 72.8           & 70.3          & 68.5          & 72.8                & 72.7          & 58.2          & 77.4          & 84.1          & 71.1    \\ \hline
\multirow{6}{*}{\rotatebox[origin=c]{90}{\scriptsize{\textit{Feat}}}} & LMMS\textsubscript{2348} (BERT)        & 68.1          & 76.3           & 75.6          & 75.1          & 77.0                & -             & -             & -             & -             & 75.4    \\
                                                                      & SemCor+WNGC, hypernyms (single)        & -             & -              & -             & -             & -                   & -             & -             & -             & -             & 77.1    \\
                                                                      & SemCor+WNGC, hypernyms (ensemble)      & 73.4          & 79.7           & \textbf{77.8} & 78.7          & 82.6                & 81.4          & 68.7          & \textbf{83.7} & 85.5          & 79.0    \\
                                                                      & SENSEMBERT\textsubscript{sup}          & -             & -              & -             & -             & -                   & 80.4          & -             & -             & -             & -       \\
                                                                      & \textit{BEM}\footnotemark              & \textit{74.5} & \textit{79.4}  & \textit{77.4} & \textit{79.7} & \textit{81.7}       & \textit{81.4} & \textit{68.5} & \textit{83.0} & \textit{87.9} & \textit{79.0} \\
                                                                      & \textit{EWISER\textsubscript{hyper}}\footnotemark[\value{footnote}] & \textit{75.2} & \textit{80.8} & \textit{79.0} & \textit{80.7} & \textit{81.8} & \textit{82.9} & \textit{69.4} & \textit{83.6} & \textit{87.3} & \textit{80.1}    \\ \hline
\multirow{2}{*}{\rotatebox[origin=c]{90}{\scriptsize{\textit{FT}}}}   & BERT\textsubscript{def}                & -             & 76.4           & 74.9          & 76.3          & 78.3                & 78.3          & 65.2          & 80.5          & 83.8          & 76.3    \\
                                                                      & GlossBERT (Sent-CLS-WS)                & 72.5          & 77.7           & 75.2          & 76.1          & 80.4                & 79.3          & 66.9          & 78.2          & 86.4          & 77.0    \\ \hline
\multirow{4}{*}{\rotatebox[origin=c]{90}{\scriptsize{\textit{Ours}}}} & BERT\textsubscript{base} (baseline)    & \textbf{73.6} & 79.4           & 76.8          & 77.4          & 81.5                & 80.6          & 67.9          & 82.2          & 87.3          & 78.2    \\
                                                                      & BERT\textsubscript{base} (augmented)   & \textbf{73.6} & 79.3           & 76.9          & 79.1          & 82.0                & 81.3          & 67.7          & 82.2          & 87.9          & 78.7    \\
                                                                      & BERT\textsubscript{large} (baseline)   & 73.0          & \textbf{79.9}  & 77.4          & 78.2          & 81.8                & 81.2          & \textbf{68.8} & 81.5          & \textbf{88.2} & 78.7    \\
                                                                      & BERT\textsubscript{large} (augmented)  & 72.7          & 79.8           & \textbf{77.8} & \textbf{79.7} & \textbf{84.4}       & \textbf{82.6} & 68.5          & 82.1          & 86.4          & \textbf{79.5}    \\ \hline
\end{tabular}
\caption{F1-score (\%) on the English all-words WSD benchmark datasets in \citet{raganato2017word}. The systems are grouped into 5 categories: i) knowledge-based system (\textit{KB}), i.e. the most frequent sense baseline, Lesk\textsubscript{ext}+emb \cite{basile2014enhanced} and Babelfy \cite{moro2014entity}, ii) supervised models (\textit{Sup}), i.e. IMS+emb \cite{iacobacci2016embeddings}, LSTM-LP \cite{yuan2016semi}, Bi-LSTM \cite{raganato2017neural} and HCAN \cite{luo2018leveraging}, iii) featured-based approach using contextual embeddings from BERT (\textit{Feat}), i.e. LMMS\textsubscript{2348} \cite{loureiro2019language}, SemCor+WNGC \cite{vial2019sense}, SENSEMBERT\textsubscript{sup} \cite{scarlini2020sensembert}, BEM \cite{blevins-zettlemoyer-2020-moving} and EWISER\textsubscript{hyper} \cite{bevilacqua2020breaking}, iv) fine-tuning approach using BERT (\textit{FT}), i.e. BERT\textsubscript{def} \cite{du2019using} and GlossBERT \cite{huang2019glossbert}, v) our models (\textit{Ours}).}
\label{table:result}
\end{table*}

\section{Experiments}
In this section, we introduce the datasets and experiment settings used to fine-tune BERT. We also present the evaluation results of each model and compare them against existing WSD systems.

\subsection{Datasets}
Both training and testing datasets were obtained from the unified evaluation framework for WSD \cite{raganato2017word}. Our training dataset for gloss selection consists of 2 parts: a baseline dataset with 226,036 training instances constructed from SemCor and an augmented dataset with 37,596 training instances constructed using the data augmentation method. When constructing the context-gloss pairs for the training datasets, we select a maximum of $n$ = 6 context-gloss pairs per training instance; for testing datasets, all possible candidate context-gloss pairs are considered.

The testing dataset contains 5 benchmark datasets from previous Senseval and SemEval competitions, including Senseval-2 (\textbf{SE2}), Senseval-3 (\textbf{SE3}), SemEval-07 (\textbf{SE07}), SemEval-13 (\textbf{SE13}), and SemEval-15 (\textbf{SE15}). Following \citet{huang2019glossbert} and others, we choose SemEval-07 as the development set for tuning hyperparameters.

\subsection{Experiment Settings}
% We experiment with both uncased BERT\textsubscript{base} and BERT\textsubscript{large} models. BERT\textsubscript{base} consists of 110M parameters with 12 Transformer layers, 768 hidden units and 12 self-attention heads while BERT\textsubscript{large} consists of 340M parameters with 24 Transformer layers, 1024 hidden units and 16 self-attention heads. We use the implementation from the transformers package\footnote{\url{https://github.com/huggingface/transformers}} \cite{wolf2019huggingface}. In total, we trained 4 models via fine-tuning on 2 setups:
% \begin{itemize}
%     \item {\bf BERT\textsubscript{base} / BERT\textsubscript{large} (baseline)}, in which only the baseline training dataset is used.
%     \item {\bf BERT\textsubscript{base} / BERT\textsubscript{large} (augmented)}, in which the concatenation of baseline and augmented training dataset is used.
% \end{itemize}

We experiment with both uncased BERT\textsubscript{base} and BERT\textsubscript{large} models. BERT\textsubscript{base} consists of 110M parameters with 12 Transformer layers, 768 hidden units and 12 self-attention heads while BERT\textsubscript{large} consists of 340M parameters with 24 Transformer layers, 1024 hidden units and 16 self-attention heads. We use the implementation from the transformers package \cite{wolf2019huggingface}. In total, we trained 4 models on 2 setups: (1) {\bf BERT\textsubscript{base/large} (baseline)}, using only the baseline dataset; (2) {\bf BERT\textsubscript{base/large} (augmented)}, using the concatenation of baseline and augmented dataset.

At fine-tuning, we set the initial learning rate to 2e-5 with batch size of 128 over 4 training epochs. The remaining hyperparameters are kept at the default values specified in the transformers package.

\footnotetext{For reference, we included the results from ACL2020. Since these results were not available at the time of writing this paper, we did not compare with the results in Section \ref{results}.}

\subsection{Evaluation Results}
\label{results}
We evaluate the performance of each model and report the F1-scores in Table \ref{table:result}, along with the results from other WSD systems.

All 4 of our models trained on the proposed gloss selection objective show substantial improvement over the non-ensemble systems across all benchmark datasets, which signifies the effectiveness of this task formulation\footnote{Statistically different from previously reported results (with p=0.05) under one-sided randomization test on the F1-scores in concatenated dataset.}. The addition of augmented training set further improves the performance, particularly in the noun category. It is worth noting that \citet{du2019using} and \citet{huang2019glossbert} reported slightly worse or identical results when fine-tuning on BERT\textsubscript{large}, but both of our models fine-tuned on BERT\textsubscript{large} obtain considerable better results than the BERT\textsubscript{base} counterparts. This may be partially attributed to the fact that we were using the recently released whole-word masking variant of BERT\textsubscript{large}, which was shown to have a better performance on the Multi-Genre Natural Language Inference (MultiNLI) benchmark. Although the BERT\textsubscript{large} (augmented) model has lower F1-score on the development dataset, it outperforms the ensemble system consisting of eight independent BERT\textsubscript{large} models on three testing datasets and achieves the best F1-score on the concatenation of all datasets.

To illustrate that the improvement of WSD performance comes from the gloss selection objective instead of hyperparameter settings, we fine-tune a BERT\textsubscript{base} model on the unagumented training set using the same hyperparameter settings as GlossBERT \cite{huang2019glossbert}, i.e. setting learning rate and batch size to 2e-5 and 64 respectively, and using 4 context-gloss pairs for each target word. As shown in Table \ref{table:compare}, our model fine-tuned on the proposed gloss selection objective consistently outperforms GlossBERT across all benchmark datasets under the same hyperparameter settings.

\begin{table}[ht]
\small
\centering
\begin{tabular}{l|llllll} \toprule
                         & SE07 & SE2  & SE3  & SE13 & SE15   \\ \hline
GlossBERT                & 72.5 & 77.7 & 75.2 & 76.1 & 80.4   \\
BERT\textsubscript{base} & 73.0 & 79.1 & 77.3 & 77.4 & 81.0   \\ \bottomrule
\end{tabular}
\caption{Comparison of F1-score (\%) on different benchmark datasets between GlossBERT and a BERT\textsubscript{base} model fine-tuned with gloss selection objective.}
\label{table:compare}
\end{table}

\section{Conclusion}
We proposed the gloss selection objective for supervised WSD, which formulates WSD as a relevance ranking task based on context-gloss pairs. Our models fine-tuned on this objective outperform other non-ensemble systems on five English all-words benchmark datasets. Furthermore, we demonstrate how to generate additional training data without external annotations using existing example sentences from WordNet, which provides extra performance boost and enable our single-model system to surpass the state-of-the-art ensemble system by a considerable margin on a number of benchmark datasets.

\section*{Acknowledgements}
We thank the meta-reviewer, the three anonymous reviewers and Ms.Vu Thi Ly for their insightful feedback and suggestions.

\bibliographystyle{acl_natbib}
\bibliography{emnlp2020}

\section*{Appendix}
\renewcommand{\thesubsection}{\Alph{subsection}}
% \subsection{Additional Details on Datasets}
% Details of the dataset used in our experiment is given in Table \ref{table:dataset}.
% \begin{table}[!htbp]
% \centering
% \begin{tabular}{lcc}
% Dataset            &  Type     & \#Annotations           \\ \hline\hline
% SemCor             &  Train    &  226,036                \\
% SemEval-07         &  Dev      &  455                    \\
% Senseval-2         &  Test     &  2,282                  \\
% Senseval-3         &  Test     &  1,850                  \\
% SemEval-13         &  Test     &  1,644                  \\
% SemEval-15         &  Test     &  1,022                                       
% \end{tabular}
% \caption{Number of annotations in each dataset}
% \label{table:dataset}
% \end{table}

\subsection{Additional Details on Experiment Settings}
All models are trained using a single Nvidia Tesla K40 GPU with 12 GB of memory. Gradient accumulation is used to accommodate large batch size.

For hyperparameters search, we manually tune for the optimal hyperparameter combinations using the following candidate values:
\begin{itemize}
    \item \textbf{BERT variant}: \{cased, uncased\}
    \item \textbf{Maximum number of glosses per context}: \{4, 6\}
    \item \textbf{Batch size}: \{32, 64, 128\}
    \item \textbf{Initial learning rate}: \{2e-5, 3e-5, 5e-5\}
    \item \textbf{Warm-up steps}: \{0, 0.1 * total steps\}
\end{itemize}

At testing stage, model checkpoints with the highest F1 score on the development dataset, i.e. SemEval-07, evaluated at every 1000 steps over 4 training epochs, are selected for evaluation on the testing dataset. We use the scoring script downloaded from \url{http://lcl.uniroma1.it/wsdeval/home}.

\end{document}

\section{Appendix}
\subsection{Additional Details on Datasets}
Details of the dataset used in our experiment is given in Table \ref{table:dataset}.
\begin{table}[!htbp]
\centering
\begin{tabular}{lcc}
Dataset            &  Type     & \#Annotations           \\ \hline\hline
SemCor             &  Train    &  226,036                \\
SemEval-07         &  Dev      &  455                    \\
Senseval-2         &  Test     &  2,282                  \\
Senseval-3         &  Test     &  1,850                  \\
SemEval-13         &  Test     &  1,644                  \\
SemEval-15         &  Test     &  1,022                                       
\end{tabular}
\caption{Number of annotations in each dataset}
\label{table:dataset}
\end{table}

\subsection{Additional Details on Experiment Settings}
All models are trained using a single Nvidia Tesla K40 GPU with 12 GB of memory. Gradient accumulation is used to accommodate large batch size.

For hyperparameters search, we manually tune for the optimal hyperparameter combinations using the following candidate values:
\begin{itemize}
    \item \textbf{BERT variant}: \{cased, uncased\}
    \item \textbf{Maximum number of glosses per context}: \{4, 6\}
    \item \textbf{Batch size}: \{32, 64, 128\}
    \item \textbf{Initial learning rate}: \{2e-5, 3e-5, 5e-5\}
    \item \textbf{Warm-up steps}: \{0, 0.1 * total steps\}
\end{itemize}

At testing stage, model checkpoints with the highest F1 score on the development dataset, i.e. SemEval-07, evaluated at every 1000 steps over 4 training epochs, are selected for evaluation on the testing dataset. We use the scoring script downloaded from \url{http://lcl.uniroma1.it/wsdeval/home}.